\documentclass[conference,onecolumn]{IEEEtran}
\IEEEoverridecommandlockouts

\usepackage{cite}
\usepackage{amsmath,amssymb,amsfonts}
\usepackage{algorithmic}
\usepackage{graphicx,subfigure}
\usepackage{textcomp}
\usepackage{xcolor}
\def\BibTeX{{\rm B\kern-.05em{\sc i\kern-.025em b}\kern-.08em
    T\kern-.1667em\lower.7ex\hbox{E}\kern-.125emX}}
\begin{document}

\title{Machine Learning for Real-Time Anomaly Detection in Optical Networks
}

\author{\IEEEauthorblockN{Sadananda Behera, Tania Panayiotou, Georgios Ellinas}
\IEEEauthorblockA{\textit{KIOS Research and Innovation
   Center of Excellence,} \\ \textit{Department of Electrical and
   Computer Engineering}, \\ \textit{University of Cyprus, Nicosia, Cyprus},  \\
   \{behera.sadananda, panayiotou.tania, gellinas\}@ucy.ac.cy
}
}

\maketitle

\begin{abstract}

This work proposes a real-time anomaly detection scheme that leverages the multi-step ahead prediction capabilities of encoder-decoder (ED) deep learning models with recurrent units. Specifically, an encoder-decoder is used to model soft-failure evolution over a long future horizon (i.e., for several days ahead) by analyzing past quality-of-transmission (QoT) observations. This information is subsequently used for real-time anomaly detection (e.g., of attack incidents), as the knowledge of how the QoT is expected to evolve allows capturing unexpected network behavior. Specifically, for anomaly detection, a statistical hypothesis testing scheme is used, alleviating the limitations of supervised (SL) and unsupervised learning (UL) schemes, usually applied for this purpose. Indicatively, the proposed scheme eliminates the need for labeled anomalies, required when SL is applied, and the need for on-line analyzing entire datasets to identify abnormal instances (i.e., UL). Overall, it is shown that by utilizing QoT evolution information, the proposed approach can effectively detect abnormal deviations in real-time. 
Importantly, it is shown that the information concerning soft-failure evolution (i.e., QoT predictions) is essential to accurately detect anomalies.   
\end{abstract}

\section{Introduction}
Effective anomaly detection schemes in optical networks are necessary to allow for repair actions to be taken before hard-failure occurs~\cite{chen2022cooperative} and to prevent the undesired consequences of malicious attacks. Hard-failures lead to service disruptions, and they can be caused by unexpected events such as fiber cuts~\cite{8320602} and certain types of physical layer attacks. Hard-failures may be also related to soft-failures eventually turning into hard-failures (e.g., gradual degradation of performance due to device ageing). While this type of soft-failure can be modeled towards predicting and preventing the event of a hard-failure~\cite{Wang:17}, there are other types of soft-failures that are not predictable (e.g., certain types of attacks such as jamming and eavesdropping) but they may eventually affect the normal operation of the network~\cite{skorin2016physical}. Thus, the detection of both types of soft-failures (i.e., predicted/unpredicted network anomalies) is of critical importance for the seamless operation of the network.     

Predictable types of soft-failures were previously addressed in~\cite{10000690}, by effectively modeling soft-failure evolution to identify, well in advance, the time that a hard-failure is expected to occur. Soft-failure evolution modeling was based on the capabilities of an encoder-decoder (ED) model with recurrent units~\cite{cho-etal-2014-learning}, capable of capturing the time-varying long-range dependencies and non-linearities of quality-of-transmission (QoT), when designed with long short-term memory (LSTM) cells~\cite{10000690}. In this work, soft-failure evolution modeling is leveraged to additionally address the unpredictable types of soft-failures. Specifically, a real-time anomaly detection scheme is proposed, that is specifically designed to detect unexpected signal deviations (i.e., anomalies) by leveraging the expected soft-failure evolution information.

Specifically, a statistical hypothesis testing scheme is proposed that is based on leveraging information regarding the predictable types of soft-failures. The insight is that the knowledge of how the predictable soft-failures are expected to evolve over-time, may contribute towards effectively capturing unexpected network behavior. In fact, it is shown that when the latter information is leveraged in the statistical hypothesis testing scheme, anomaly detection accuracy is increased, compared to the case where this information is not considered; an indicator of the importance of considering expected (i.e., future) soft-failure information to detect unexpected anomalies.    

\subsection{Related Work}
There have been many previous studies on real-time anomaly detection in optical networks. One common approach is to use SL techniques \cite{malhotra2015long,salman2017machine} (e.g., random forests, auto-encoders) to train a model on a labeled dataset of normal and abnormal network incidents. However, these methods require a large amount of labeled data that may not be available for real implementations.   

To overcome the limitations of supervised learning (SL), unsupervised learning (UL) techniques (e.g., clustering, density estimation), were instead applied to identify abnormal instances in unlabeled datasets \cite{lin2015cann,chen2022cooperative,abdelli2022machine}. Even though, in general, UL does not require labeled data, it still may require large datasets to effectively perform anomaly detection. These datasets need to also be continuously processed on-line, in order to perform anomaly detection. Hence, UL constitutes a computationally inefficient approach~\cite{Xu15}.  Nevertheless, in~\cite{chen2022cooperative,abdelli2022machine} UL is applied once to automatically label the dataset to normal and abnormal incidents, aiming to solve some limitations of SL.

The major limitation of existing SL- and UL-based anomaly detection schemes is, however, that they do not consider how predictable soft-failures evolve over time, leading to the detection of premature (i.e., more frequent than necessary) repair actions~\cite{10000690}. In this work, the evolution of predictable soft-failures is taken into account to allow a network operator to differentiate between the soft-failures that need to be urgently addressed (e.g., attacks, a soft-failure that is close to turning into a hard-failure), and the soft-failures that are not expected to cause severe malfunctions any time soon.     

\subsection{Contribution}
This work proposes a statistical hypothesis testing scheme that leverages predictable soft-failure evolution over a long future horizon, to effectively detect unpredictable types of soft-failures. 
As such, it eliminates the need for large labeled datasets (as required by SL), as well as the need for on-line analyzing entire datasets to identify abnormal instances (as required by UL). It is demonstrated that the information of predictable soft-failure evolution is essential as it enhances the accuracy of the proposed scheme. Overall, the proposed approach succeeds in differentiating between predictable and unpredictable soft-failures, allowing the network operator to initiate appropriate repair actions only when it is necessary. It is worth mentioning that hypothesis testing was previously applied for other types of networks and applications. Indicatively, in~\cite{herodotou2014scalable}, hypothesis testing was used to remove background noise from training datasets that is present in large-scale network monitoring systems. To the best of our knowledge this is the first time that hypothesis testing is utilized for anomaly detection in optical networks.

\section{Approach Overview}\label{overview}
\begin{figure}[htbp]
\subfigure[]{\includegraphics[scale =0.32]{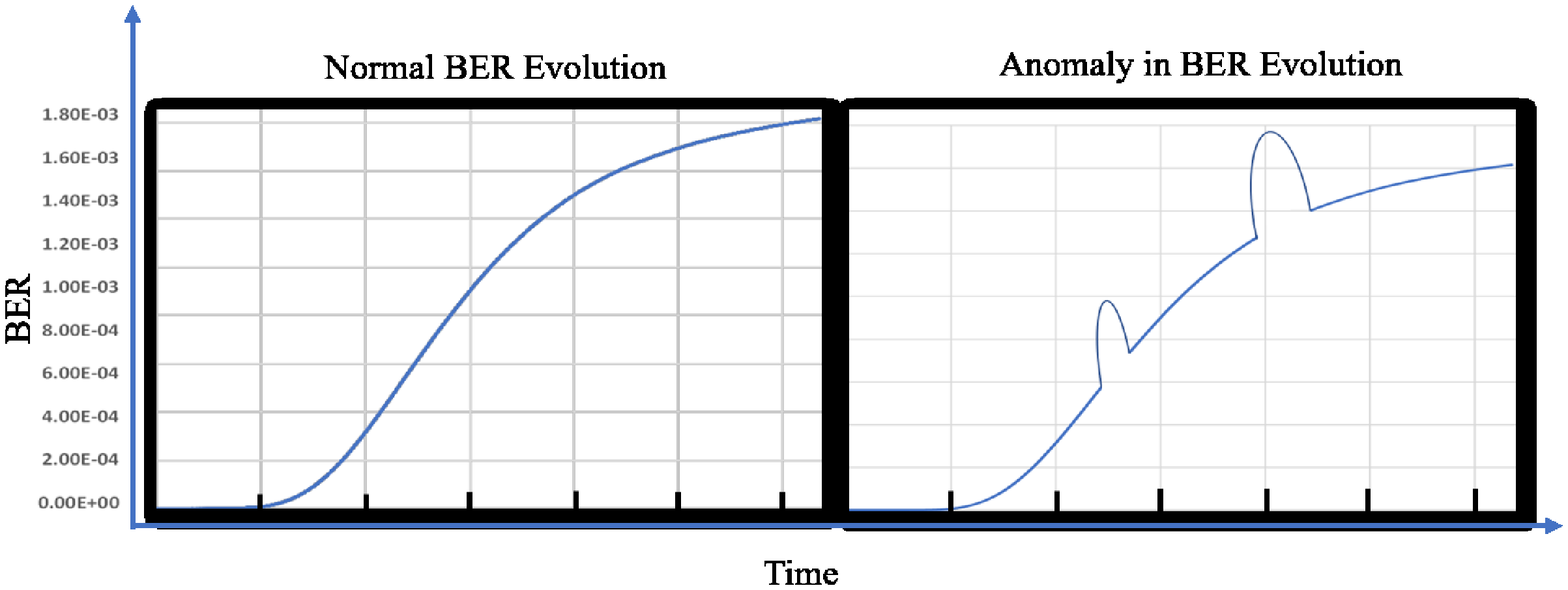}\label{ber_anomaly}} 
\subfigure[]{\includegraphics[trim={0cm 6cm 0cm 0cm}, clip,scale =0.3]{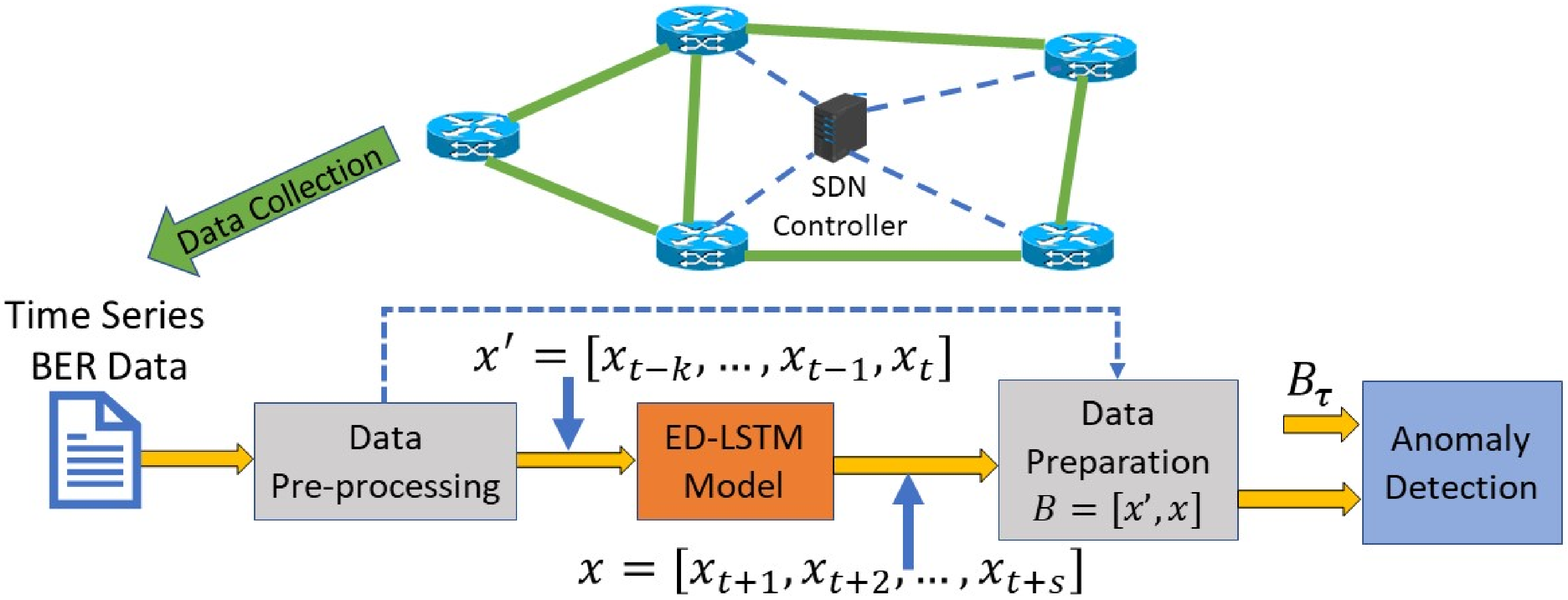}\label{overview}} 
\vspace{-.2in}
\caption{Methodology overview; (a) Example of normal BER evolution and anomaly in BER evolution  (b) Proposed anomaly detection scheme.}
\label{Overview}
\end{figure}

Figure~\ref{ber_anomaly} depicts a scenario where the evolution of bit error rate (BER) over time distinguishes between normal behavior and the presence of anomalies, while Fig.~\ref{overview} provides an overview of the proposed anomaly detection scheme, which consists of two phases; that is, soft-failure evolution modeling, followed by real-time anomaly detection. First, an ED-LSTM model is trained/tested over a dataset that contains sequences of BER information monitored at the coherent receivers of an optical node collected using software-defined networking (SDN) controllers. Specifically, the ED-LSTM learns a nonlinear function $f(.)$, that, given the past and present observations $x'=[{x}_{t-k}, \cdots, { x}_{t-1},{x}_t]$, accurately predicts future BER observations 
$x=[{x}_{t+1}, { x}_{t+2}, \cdots, {x}_{t+s}]$, where $s$ is the future time step, $k$ is the number of past and present observations, and $t$ is the current time instant. After model training and testing, the ED-LSTM model can be used on-line over unseen (new) $x'$ sequences to: (i) predict the time that a soft-failure is expected to turn into a hard-failure~\cite{10000690}, and (ii) perform detection of unexpected anomalies. Note that on-line anomaly detection also serves to ``clean'' (i.e., pre-process) new datasets, that are subsequently used for updating (i.e., re-training) the ED-LSTM model in the presence of non-stationary BER evolution. For anomaly detection, that this work focuses on, a hypothesis testing scheme is proposed that combines statistical metrics with the on-line predictions of the ED-LSTM model to identify in real-time deviations in the monitored BER data.

\subsection{Anomaly Detection Scheme}
Given a sequence of BER samples $B = [x', x]$, ultimately forming a BER distribution over a specific time interval, the proposed approach aims to determine if a BER observation $B_{\tau}$ deviates from its historical and future (i.e., $x',x$) BER values; that is, $B_{\tau}$ is a BER value monitored within the time interval spanned by $x$. In other words, it aims to detect whether $B_{\tau}$ belongs to the distribution $B$ or not. If the sample does not belong to $B$ (i.e., a deviation exists), then an anomaly is detected, subsequently triggering the necessary actions (i.e., identification of anomaly, repair action, etc.).  
To achieve this, a series of Student's $\mathsf{t}$-tests are used to determine whether the addition of the $B_{\tau}$ value in the distribution $B$ introduces a statistically significant change in the sequence of BER values. Specifically, the hypothesis testing scheme is formulated as follows: 
Let $n_{\tau}$, $\mu_{\tau}$, and $\sigma^{2}_{\tau}$ represent the size, mean, and variance of $B$, respectively. An anomaly is detected when
\vspace{-0.05in}
 \begin{equation}
     B_{\tau} > \theta_{
     \tau
     } . \mu_{
     \tau
     } + \mathsf{t}_{\alpha,n_{\tau}-1}.\sqrt{\frac{\sigma^{2}_{
     \tau
     }}{n_{\tau}}},
     \label{eq1}
 \end{equation}
where $\theta_{\tau}$ is a scaling parameter, $\mathsf{t}_{\alpha,n_{\tau}-1}$ is the $\mathsf{t}$ value obtained from the $\mathsf{t}$-distribution using significance level $\alpha=99\%$, and $n_{\tau}-1$ is the degree of freedom. As such, $B_{\tau}$ is considered anomalous when it is deemed significantly different from $B$.

\subsection{Benchmark Anomaly Detection Scheme}
As a benchmark, the hypothesis testing scheme previously described is applied (i.e., Eq.~\eqref{eq1}), with the difference that $B = [x']$. Hence, during the hypothesis testing it is assumed that ED-LSTM predictions $x$ are not available. The purpose is to examine whether future soft-failure evolution constitutes an important information for anomaly detection or not.

\section{Performance Evaluation}\label{evaluation}
\begin{figure}[htbp]
\begin{center}
\subfigure[]{\includegraphics[trim={2cm 1cm 6cm 3cm}, clip,height=.23\textwidth, width=.33\textwidth]{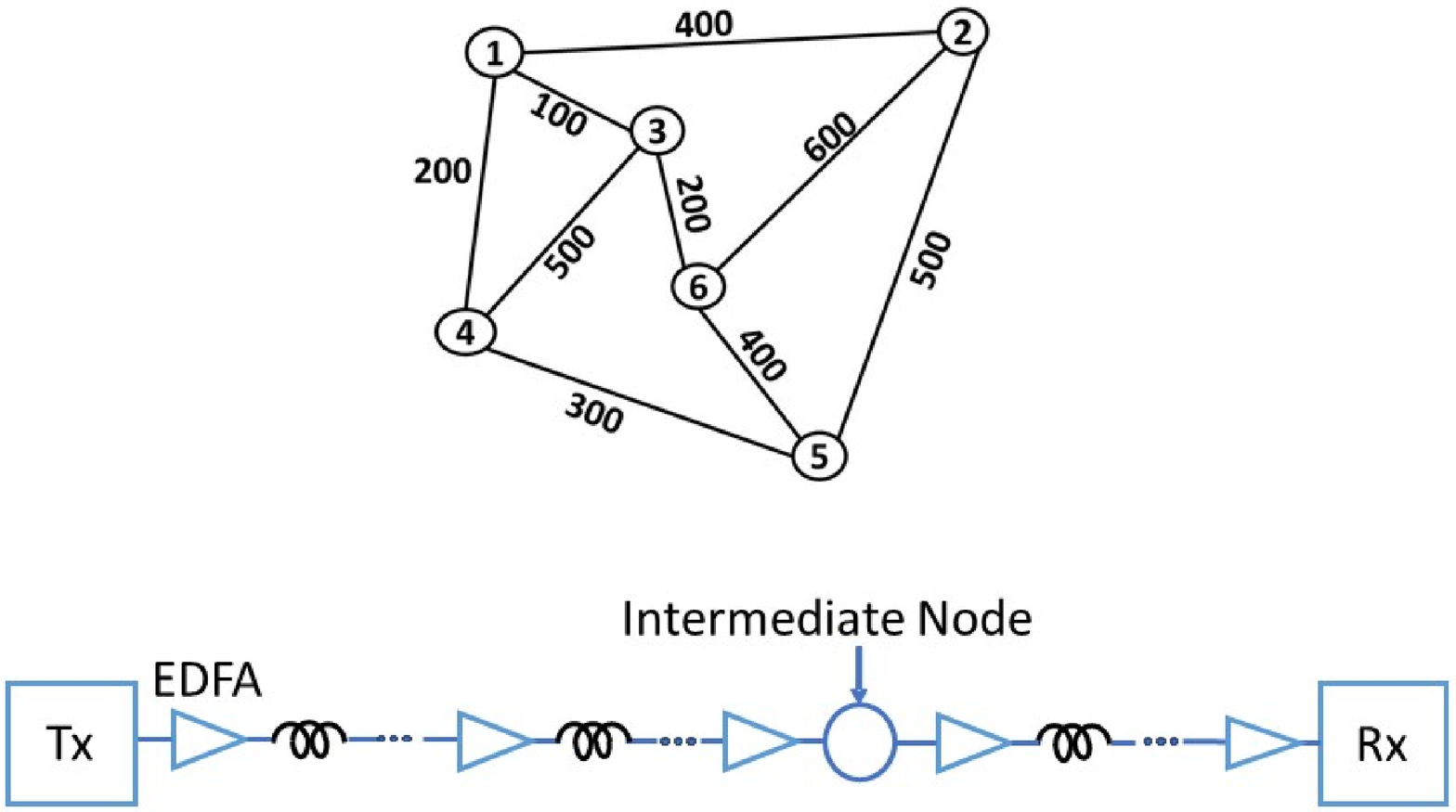}\label{testbed}} 
\subfigure[]{\includegraphics[height=.25\textwidth, width=.32\textwidth]{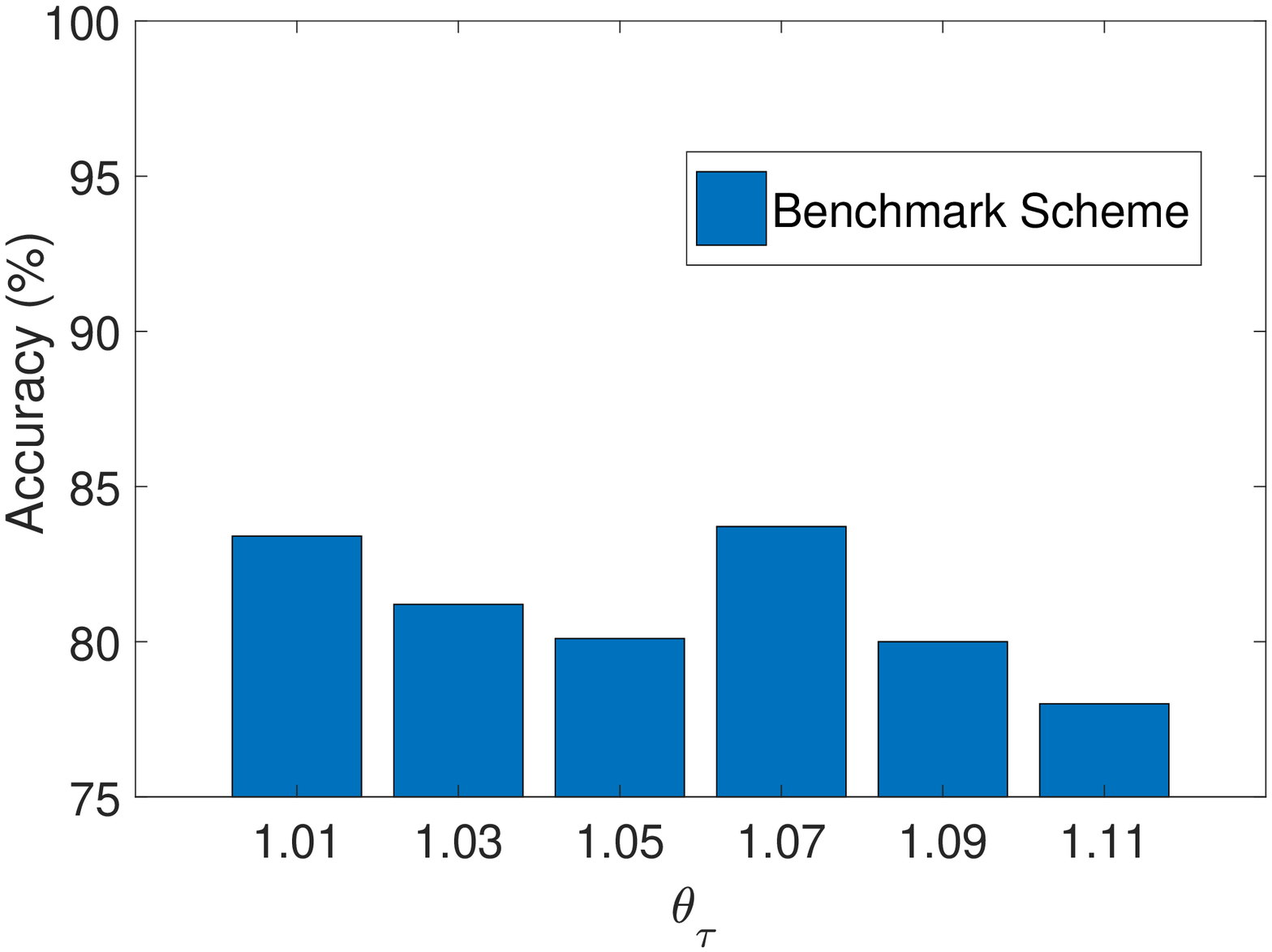}\label{benchmark}} 
\subfigure[]{\includegraphics[height=.25\textwidth, width=.32\textwidth]{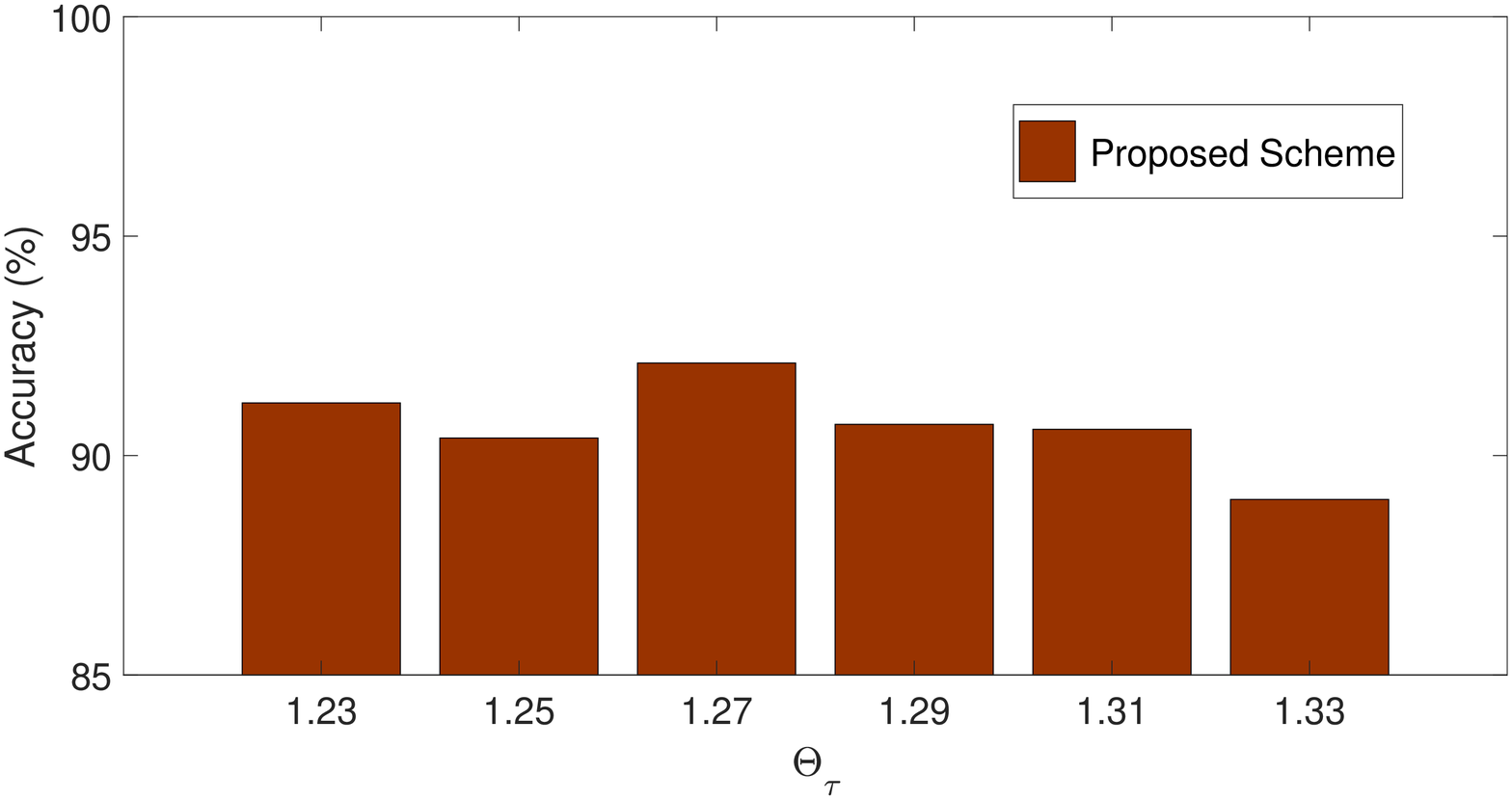}\label{proposed}} 
\vspace{-.3cm}
\caption{(a) 6-node topology and generic simulation setup; Accuracy of the anomaly detection for both the benchmark (b) and proposed (c) schemes for various values of the scaling parameter $\theta_\tau$.}
\label{accy}
\end{center}
\end{figure}

For modeling predictable soft-failure evolution, the ED-LSTM architecture, ED-LSTM hyperparameters, physical layer model (PLM) for elastic optical networks~\cite{behera2019impairment}, and dataset analytically described in~\cite{10000690} are utilized. In brief, the dataset is generated synthetically by sequentially inducing soft-failures (i.e., erbium-doped fiber amplifier (EDFA) gain degradation) into the PLM utilizing  the generic simulation setup for a $6$-node topology as shown in Fig.~\ref{testbed}. Specifically, EDFA degradation follows the Weibull distribution and is induced over a single lightpath  of the $6$-node topology (i.e., lightpath $(6,4)$ traversing links $(6,5)$ and $(5,4)$), with the EDFAs spaced $100$ km apart. The reader should note that this simulation setup constitutes just an instance of the network state, used to demonstrate the prediction capabilities of the ED-LSTM model for anomaly detection. A similar methodology can be used for other network instances
as well, by partitioning the network state into groups of lightpaths. Nevertheless, sequences $x',x$ are created following the sliding window approach with $k = 50$ and $s = 70$, with each BER value in $x',x$ sampled every $1.5$ hours. Training is performed over a set of $5472$ sequences, and testing over the following $609$ sequences, demonstrating an accuracy of $1.26 \times 10^{-7}$. Overall, in~\cite{10000690} it was shown that the trained ED-LSTM can accurately predict the event of a hard-failure up to $4$ days ahead of time, allowing for the necessary repair actions to take place.  

The proposed anomaly detection scheme is evaluated over the predictions of the test sequences. Specifically, unexpected anomalies are created over the period spanning the $609$ sequences (i.e.,  the interval $h$) by creating random BER deviations. These deviations are created by randomly increasing the expected BER of samples that belong to interval $h$ by $7-15\%$ (an expected BER behavior upon most types of known anomalies)~\cite{abdelli2022machine}. In total, $12$ anomalies are randomly induced (i.e., a typical percentage of abnormal instances in the dataset~\cite{chen2022cooperative}) and anomaly detection is performed every $5$ minutes (i.e., $B_{\tau}$ is monitored and tested every $5$ minutes). 

Figures~\ref{benchmark} and \ref{proposed} show the accuracy of the anomaly detection for both the benchmark and proposed schemes for various values of the scaling parameter  $\theta_\tau$. Specifically, Figs.~\ref{benchmark} and \ref{proposed} show the averaged accuracy (over $1000$ simulations) achieved within the class of anomalous instances (i.e., over the $12$ anomalies). It should be noted, that the range of $\theta_\tau$ values illustrated in Figs.~\ref{benchmark} and \ref{proposed} corresponds to the range of values where false positives are close to zero; a critical criterion that is set towards the exploration of the best $\theta_\tau$ value to ultimately prevent unnecessary repair actions (i.e., false alarms). Thus, the range of $\theta_\tau$ values utilized results in an accuracy of $100\%$ within the normal $B_{\tau}$ instances for both proposed and benchmark schemes. Overall, these $\theta_\tau$ values are randomly generated and the optimal range is found empirically ($\theta_\tau \in [1.01 \ \ 1.11]$ for the benchmark and $\theta_\tau \in [1.23 \ \ 1.33]$ for the proposed scheme) as shown in Figs.~\ref{benchmark}-\ref{proposed}. 

According to the results, the proposed scheme, utilizing the ED-LSTM predictions, outperforms the benchmark scheme for all scaling parameters examined (i.e., within their corresponding optimal range). 
Specifically, the proposed method achieved the highest accuracy of $92.11\%$ at $\theta_\tau=1.27$, while the benchmark achieved the highest accuracy of $83.71\%$ at $\theta_\tau=1.07$. Hence, the proposed approach results in an improvement in accuracy that is shown to be up to $8.4\%$ for this simulation setting, reaffirming our insight that predictions can aid in detecting anomalies.

To further assess the effectiveness of the proposed anomaly detection model, additional metrics are examined, namely the Precision $(P)$, Recall $(R)$, and $F$-measure as given below \cite{jia2019anomaly}:
\begin{equation}
   P=\frac{t_p}{t_p+f_p};  R=\frac{t_p}{t_p+f_n}; F=2\times\frac{P*R}{P+R}
\end{equation}
where, $t_p$, $f_p$, $f_n$ denote the number of true positives, false positives, and false negatives, respectively. 

\begin{figure}[htbp]
\begin{center}
\subfigure[]{\includegraphics[height=.23\textwidth, width=.45\textwidth]{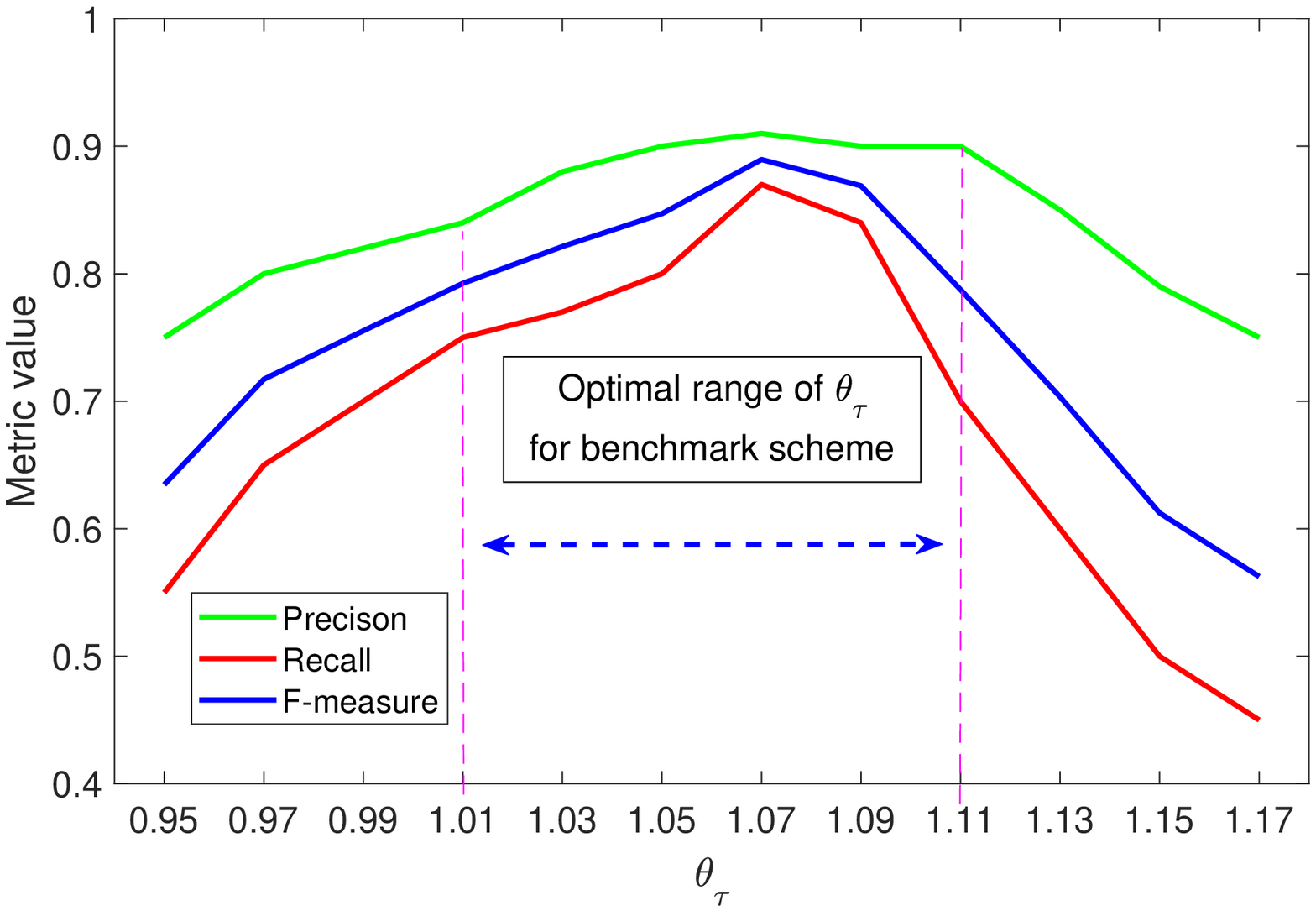}\label{benchmark_prf}} 
\subfigure[]{\includegraphics[height=.23\textwidth, width=.45\textwidth]{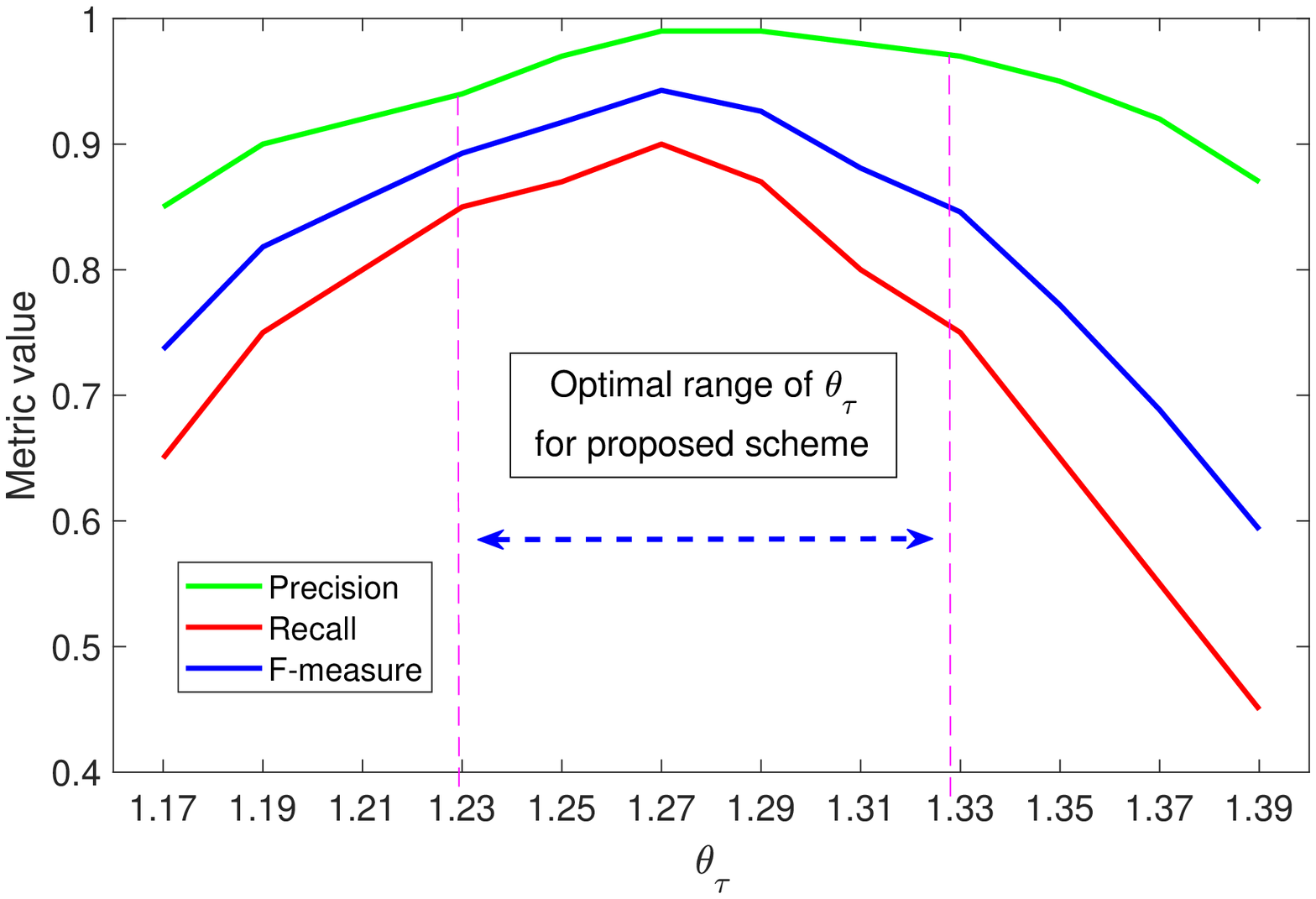}\label{proposed_prf}} 
\vspace{-.3cm}
\caption{Performance of anomaly detection metrics corresponding to different scaling parameter ($\theta_\tau$) range  for (a) Benchmark scheme (b) Proposed scheme.}
\label{prf}
\end{center}
\end{figure}

According to Fig.~\ref{prf}, the Precision, Recall, and $F$-measure metrics are evaluated for various values of $\theta_\tau$ using both the benchmark and proposed anomaly detection schemes. The obtained results indicate that the optimal range of $\theta_\tau$ for both schemes is the same as the one reported in Figs.~\ref{benchmark}-\ref{proposed}. The highest $F$-measure value, which represents the best balance between Precision and Recall, is observed when $\theta_\tau$ is set to $1.07$ for the benchmark scheme (Fig.~\ref{benchmark_prf}), and $1.23$ for the proposed scheme (Fig.~\ref{proposed_prf}). Therefore, these values of $\theta_\tau$ are recommended as the optimal settings for the benchmark and proposed schemes.

\section{Conclusion}\label{con}
An anomaly detection scheme is proposed that is based on statistical hypothesis testing and the capabilities of an ED-LSTM model on accurately modeling predictable soft-failure evolution. This scheme is effective in accurately detecting anomalies in real-time (up to $92.11\%$ accuracy), outperforming (up $8.4\%$ improvement) an anomaly detection scheme that does not consider the expected evolution of soft-failures; an indicator of the importance of considering this information in anomaly detection. The accuracy of the proposed scheme can be further improved by considering additional statistical properties and/or by (automatically) fine tuning scaling parameter $\theta_\tau$. Finally, soft-failure evolution information can be used in conjunction with other techniques (e.g., UL) to enhance detection accuracy.   

\section*{Acknowledgements}
This work has been supported by the European Union’s
Horizon 2020 research and innovation programme under
grant agreement No. 739551 (KIOS CoE - TEAMING) and
from the Republic of Cyprus through the Deputy Ministry of Research, Innovation and Digital Policy.


\end{document}